\documentclass{svjour3}                     % onecolumn (standard format)
\smartqed  % flush right qed marks, e.g. at end of proof

\usepackage{helvet}
\usepackage{array}
\usepackage{courier}
\usepackage[latin9]{inputenc}
\usepackage{amsmath}
\usepackage{amssymb}
\usepackage{graphicx}
\usepackage{epstopdf}
\usepackage{esint}
\usepackage{paralist}
\usepackage{enumerate}
\usepackage{cite}

\makeatletter
\DeclareMathOperator*{\argmin}{arg\,min}

\usepackage{algorithm, caption}

\usepackage{amsfonts}

\usepackage{amsthm}
\usepackage{algcompatible}
\usepackage{multirow}
\usepackage{color}
\usepackage{breakcites}

\newtheorem{assumption}{Assumption}
\newcommand\independent{\protect\mathpalette{\protect\independenT}{\perp}}
\def\independenT#1#2{\mathrel{\rlap{$#1#2$}\mkern2mu{#1#2}}}

\ifodd 1
\newcommand{\com}[1]{\textbf{\color{red}(COMMENT: #1)}} 
\newcommand{\rev}[1]{{\color{black}#1}}
\else
\newcommand{\com}[1]{}
\newcommand{\rev}[1]{#1}
\fi

\begin{document}

\title{Constructing Effective Personalized Policies Using Counterfactual Inference from Biased Data Sets with Many Features%\thanks{Grants or other notes
}

\titlerunning{Constructing Policies Using Counterfactual Inference with Many Features}        % if too long for running head

\author{Onur Atan        \and
        William R. Zame \and
        Qiaojun Feng \and
        Mihaela van der Schaar
         %etc.
}


\institute{Onur Atan \at
              University of California, Los Angeles \\
              \email{oatan@ucla.edu}           %  \\
           \and
           William R. Zame \at
              University of California, Los Angeles and Nuffield College, Oxford University\\
              \email{zame@econ.ucla.edu}  
             \and
           Qiaojun Feng \at
              Tsinghua University \\
              \email{fqj13@mails.tsinghua.edu.cn} 
             \and
           Mihaela van der Schaar \at
              Oxford-Man Institute, Oxford University and University of California, Los Angeles \\
              \email{mihaela.vanderschaar@eng.ox.ac.uk}  
}

\date{Received: -- / Accepted: --}

\maketitle

\begin{abstract}
This paper proposes a novel approach for constructing effective personalized policies when the observed data lacks counter-factual information, is biased and possesses many features.  The approach is applicable in a wide variety of settings from healthcare to advertising to education to finance.  These settings have in common that the decision maker can observe, for each previous instance, an array of features of the instance, the action taken in that instance, and the reward realized -- but not the rewards of actions that were not taken: the counterfactual information.  Learning in such settings is made even more difficult because the observed data is typically biased by the existing policy (that generated the data) and because the array of features that might affect the reward in a particular instance -- and hence should be taken into account in deciding on an action in each particular instance -- is often vast.  The approach presented here estimates propensity scores for the observed data, infers counterfactuals, identifies a (relatively small) number of features that are (most) relevant for each possible action and instance, and prescribes a policy to be followed.  Comparison of the proposed algorithm against  state-of-art algorithms on actual datasets demonstrates that the proposed algorithm achieves a significant improvement in performance.
\end{abstract}

\section{Introduction}
\label{intro}
The ``best'' treatment for the current patient must be learned from the treatment(s) of previous patients.  However, no two patients are ever {\em exactly} alike, so the learning process must involve learning the ways in which the current patient {\em is} alike to previous patients -- i.e., has the same or similar features -- and which of those features are {\em relevant} to the treatment(s) under consideration.  This already complicated learning process is further complicated because the history of previous patients records only outcomes actually experienced from treatments actually received -- not the outcomes that would have been experienced from alternative treatments -- the {\em counterfactuals}. And this learning process is complicated still further because the treatments received by previous patients were (typically) chosen according to some protocol that might or might not be known but was almost surely not random -- so the observed data is {\em biased}.  

The same complications arise in many other settings. Which mode of advertisement would be most effective for a given product? Which materials would best promote learning/performance for a given student? Which investment strategy would yield higher returns or lower risk in a particular macroeconomic environment? As in the medical setting, choosing the ''best'' policy in these settings (and in others too numerous to mention) requires learning which features of each context are relevant for the decision/action at hand and learning about the consequences of decisions/actions not taken in previous contexts -- the {\em counterfactuals}; such learning is especially complicated because the observed data may be biased (because it was created by an existing -- perhaps less effective -- policy) and because each observed instance and action may be informed by a vast array of features. (Counterfactuals are seldom seen in observed data. One possible way to obtain counterfactual information would be to conduct controlled experiments -- but in many contexts, experimentation will be impractical or even impossible. Absent controlled experiments, counterfactuals must be {\em inferred}.) 

This paper proposes a novel approach to addressing such problems. \rev{ We construct an algorithm that learns a nonlinear  policy to recommend an action for each (new) instance. During the training phase, our algorithm  learns the action-dependent relevant features and then uses a feedforward neural network to optimize a nonlinear stochastic policy the output of which is a probability distribution over the actions given the relevant features. When we apply the trained algorithm to a new instance, we choose the action which has the highest probability. } In the settings mentioned above our algorithm constructs: (in the medical context) a personalized plan of patient treatment; (in the advertising context) a product-specific plan of advertisement; (in the educational context) a student-specific plan of instruction; (in the financial context) a situationally-specific investment strategy. We use actual data to demonstrate that our algorithm is significantly superior to existing state-of-the-art algorithms.  We emphasize that our methods and the algorithms we develop are widely applicable to an enormous range of settings, from healthcare to advertisement to education to finance to recommender systems to smart cities.  (See \cite{athey2015machine}, ~\cite{hoiles2016bounded} and \cite{bottou2013counterfactual}, for just a few examples.)

As we have noted, our methods and algorithms apply in many settings, each of which comes with specific features, actions and rewards. In the medical context, typical features are items available in the electronic health record (laboratory tests, previous diagnoses, demographic information, etc.), typical actions are choices of treatments (perhaps including no treatment at all), and typical rewards are recovery rates or 5-year survival rates. In the advertising context, typical features are the characteristics of a particular website and user, typical actions are the placements of an advertisement on a webpage, and typical rewards are click-rates. In the educational context, typical features are previous coursework and grades, typical actions are materials presented or subsequent courses taken, and typical rewards are final grades or graduation rates. In the financial context, typical features are aspects of the macroeconomic environment (interest rates, stock market information, etc.), typical actions are the timing of particular investment choices, and typical rewards are returns on investment.  

For a simple but striking example from the medical context, consider the problem of choosing the best treatment for a patient with kidney stones. Such patients are usually classified by the size of the stones: small or large; the most common treatments are Open Surgery and Percutaneous Nephrolithotomy. Table 1 summarizes the results. Note that Open Surgery performs better than Percutaneous Nephrolithotomy for patients with small stones {\em and} for patients with large stones but Percutaneous Nephrolithotomy performs better overall.\footnote{This is a particular instance of Simpson's Paradox.}  Of course this would be impossible if the subpopulations that received the two treatments were identical -- but they were not.  And in fact we do not know the policy that created these subpopulations by assigning patients to treatments. We do know that patients are distinguished by a vast array of features in addition to the size of stones -- age, gender, weight, kidney function tests, etc. -- but we do not know which of these features is relevant. And of course we know the result of the treatment actually received by each patient -- but we do not know what the result of the alternative treatment would have been (the counterfactual).  

\begin{table*}[t]
\normalsize
\centering
  \caption{Success rates of two treatments for kidney stones~\cite{bottou2013counterfactual}}
    \label{table:illustrate}
    \begin{tabular}{|p{30mm}|p{23mm}|p{23mm}|p{23mm}|}
    \hline
      & Overall & Small stones & Large stones \\ \hline
    Open Surgery& $78\%(273/350)$ & $\boldsymbol{93\%(81/87)}$ & $\boldsymbol{73\%(192/263)}$  \\ \hline
    Percutaneous Nephrolithotomy & $\boldsymbol{83\%(289/350)}$ & $87\%(234/270)$ & $69\%(55/80)$   \\ \hline
    \end{tabular}
\end{table*}

Three more points should be emphasized. Although Table 1 shows only two actions, in fact there are a number of other possible actions for kidney stones: they could be treated using any of a number of different medications, they could be treated by ultrasound, or they could not be treated at all. This is important for several reasons. The first is that a number of existing methods assume that there are only two actions (corresponding to treat or not-treat); but as this example illustrates, in many contexts (and in the medical context in particular), it is {\em typically} the case that there are {\em many} actions, not just two -- and, as the papers themselves note,  these  methods simply do not work when there are more than two actions; see~\cite{johansson2016learning}.  The second is that the features that are relevant for predicting the success of a particular action typically depend on the action: different features will be found to be relevant for different actions. (The treatment of breast cancer, as discussed in \cite{yoon2016discovery}, illustrates this point well.  The issue is not simply whether or not to apply a regime of chemotherapy, but {\em which} regime of chemotherapy to apply. Indeed, there are at least six widely used regimes of chemotherapy to treat breast cancer, and the features that are relevant for predicting success of a given regime are different for different regimes.) The third is that we go much further than the existing literature by allowing for {\em nonlinear} policies. To do this, we use a feedforward neural network, rather than relying on familiar algorithms such as POEM ~\cite{swaminathan2015batch}. To determine the best treatment, the bias in creating the populations, the features that are relevant {\em for each action} and the  policy must all be {\em learned}.  Our methods are adequate to this task.

The remainder of the paper is organized as follows. In Section~\ref{sec:related}, we describe some related work and highlight the differences with respect to our work. In Section~\ref{sec:data}, we describe the observational data on which our algorithm operates.  In Section~4, we begin with an informal overview, then give the formal description of our algorithm (including substantial discussion).  Section~5 gives the pseudo-code for the algorithm.  Some extensions are discussed in Section~\ref{sec:ext}. In Section~\ref{sec:num}, we demonstrate the performance of our algorithm on a variety of real datasets.   Section 8 concludes.  Proofs are in the Appendix.

\section{Related Work}
\label{sec:related}

From a conceptual point of view, the paper most closely related to ours -- at least among recent papers -- is perhaps \cite{johansson2016learning} which treats a similar problem: learning relevance in an environment in which the counterfactuals are missing, data is biased and each instance may have many features. The approach taken there is somewhat different from ours in that, rather than identifying the relevant features, they transfer the features to a new representation space. (This process is referred as {\em domain adaptation}~\cite{johansson2016learning}.)  A more important difference from our work is that it assumes that there are only two actions: treat and don't treat. As we have discussed in the Introduction, the assumption of two actions is unrealistic; in most situations there will be {\em many} (possible) actions. It states explicitly that the approach taken there does not work when there are more than two actions and offers the multi-action setting as an obvious but difficult challenge.  One might think of our work as ``solving'' this challenge -- but we stress that the ``solution'' is not at all a routine extension.  Moreover, in addition to this obvious challenge, there is a more subtle -- but equally difficult -- challenge: when there are more than two actions, it will typically be the case that some features will be relevant for some actions and not for others, and -- as discussed in the Introduction -- it will be crucial to learn which features are relevant for which actions.  

From a technical point of view, our work is perhaps most closely related to~\cite{swaminathan2015batch} in that we use similar methods (IPS-estimates and empirical Bernstein inequalities) to learn counterfactuals. However, it does not treat observational data in which the bias is unknown and does not learn/identify relevant features. Another similar work on policy optimization from observational data is~\cite{strehl2010learning}. 

The work in~\cite{wager2015estimation} treats the related (but somewhat different) problem of estimating individual treatment effects. The approach there is through causal forests as developed by\cite{athey2015machine}, which are variations on the more familiar random forests.  However, the emphasis in this work is on asymptotic estimates, and in the many situations for which the number of (possibly) relevant features is large the datasets will typically not be large enough that asymptotic estimates will be of more than limited interest. There are many other works focusing on estimating treatment effects; some include ~\cite{tian2012simple, alaa2017bayesian, shalit2016estimating}.

More broadly, our work is related to methods for feature selection and counterfactual inference. The literature on feature selection can be roughly divided into categories according to the extent of supervision:  supervised feature selection~\cite{song2012feature,weston2003use}, unsupervised feature selection~\cite{dy2004feature,he2005laplacian} and semi-supervised feature selection~\cite{xu2010discriminative}. However, our work does not fall into any of these categories; instead we need to select features that are informative in determining the rewards of each action. This problem was addressed in~\cite{tekin2014discovering} but in an {\em online} Contextual Multi-Armed Bandit (CMAB) setting in which experimentation is used to learn relevant features. In the present paper, we treat the  {\em logged} CMAB setting in which experimentation is impossible and relevant features must be learned from the existing logged data. As we have already noted, there are many circumstances in which experimentation is impossible. The difference between the settings is important -- and the logged setting is much more difficult -- because in the online setting it is typically possible to {\em observe} counterfactuals, while in the current logged setting it is typically {\em not} possible to observe counterfactuals, and because in the online setting the decision-maker controls the observations so whatever bias there is in the data is known.  

With respect to learning, feature selection methods can be divided into three categories -- filter models, wrapper models, and embedded models~\cite{tang2014feature}. Our method is most similar to filter techniques in which features are ranked according to a selected criterion such as a Fisher score~\cite{duda2012pattern}, correlation based scores~\cite{song2012feature}, mutual information based scores~\cite{koller1996toward,yu2003feature,peng2005feature}, Hilbert-Schmidt Independence Criterion (HSIC)~\cite{song2012feature} and Relief and its variants \cite{kira1992practical,robnik2003theoretical}) etc., and the features having the highest ranks are labeled as relevant. However, these existing methods are developed for classification problems and they cannot easily handle datasets in which the rewards of actions not taken are missing.

The literature on counterfactual inference can be categorized into three groups: direct, inverse propensity re-weighting and doubly robust methods. The direct methods compute counterfactuals by learning a function mapping from feature-action pair to rewards~\cite{prentice1976use,wager2015estimation}. The inverse propensity re-weighting methods compute unbiased estimates by weighting the instances by their inverse propensity scores~\cite{swaminathan2015batch,joachims2016counterfactual}. The doubly robust methods compute the counterfactuals by combining direct and inverse propensity score reweighing methods to compute more robust estimates~\cite{dudik2011doubly,jiang2015doubly}. With respect to this categorization, our techniques might be view as falling into doubly robust methods.

\rev{ Our work can be seen as building on and extending the work of \cite{swaminathan2015batch,swaminathan2015self}, which learn {\em linear} stochastic policies.  We go much further by learning a {\em non-linear } stochastic policy. Our work can also be seen as an off-line variant of the on-line REINFORCE algorithm~\cite{williams1992simple}.}

\rev{We should also note two papers that were written {\em after} the current paper was originally submitted. The work of \cite{joachimsdeep}  extends the earlier work of   \cite{swaminathan2015batch,swaminathan2015self} to non-linear policies. Our own (preliminary) work~\cite{atan2018learning} propose a different approach for learning a representation function and a policy. Unlike the present paper, our more recent work uses a loss function that embodies both a policy loss (similar to, but slightly different than, the policy loss used in the present paper) {\em and} a domain loss (which quantifies the divergence between the logging policy and the uniform policy under the representation function). The advantage of these changes is that they make it possible to learn the representation function and the policy in an end-to-end fashion.}

\section{Data}\label{sec:data} 
We consider logged contextual bandit data: that is, data for which we know the features of each instance, the action taken and the reward realized in that instance -- but not the reward that would have been realized had a different action been taken. We assume that the data has been logged according to some policy {\em which we may not know, but which is not necessarily random} and so the data is {\em biased}. Each data point consists of a feature, an action and a reward. A {\em feature} is a vector $(x_1, \ldots, x_d)$ where each $x_i \in  \mathcal{X}_i$ is a {\em feature type}. The space of all feature types is $\mathcal{F} = \{1, \ldots, d\}$, the space of all features is $\mathcal{X} = \Pi_{i=1}^d \mathcal{X}_i$ and the set of {\em actions} is $\mathcal{A}$. We assume that the sets of feature types and actions are finite; we write  $b_i = |\mathcal{X}_i| $ for the cardinality of $\mathcal{X}_i$ and  $\mathcal{A} = \{1,2, \ldots, k \}$ for the set of actions. For $\boldsymbol{x} \in \mathcal{X}$ and $\mathcal{S} \subset \mathcal{F}$ we write $\boldsymbol{x}_\mathcal{S}$ for the restriction of $\boldsymbol{x}$ to $\mathcal{S}$; i.e. for the vector of feature types whose indices lie in $\mathcal{S}$. It will be convenient to abuse notation and view $\boldsymbol{x}_{\mathcal S}$ both as a vector of length $|{\mathcal S}|$ or as a vector of length $d = |{\mathcal F}|$ which is $0$ for feature types not in $\mathcal S$.  A {\em reward} is a real number; we normalize so that rewards lie in the interval  $[0,1]$.  In some cases, the reward will be either $1$ or $0$ (success or failure; good or bad outcome); in other cases the reward may be interpreted as the probability of a success or failure (good or bad outcome).

We are given a data set 
$$
\mathcal{D}^n = \{(\boldsymbol{X}_1, A_1, R^{\text{obs}}_1), \ldots,  (\boldsymbol{X}_n, A_n, R^{\text{obs}}_n) \}
$$  
We assume that the $j$-th instance/data point $(\boldsymbol{X}_j, A_j, R^{\text{obs}}_j)$ is generated according to the following process:
\begin{enumerate}
\item The instance is described by a feature vector $\boldsymbol{X}_j$ that arrives according to the fixed but unknown distribution $\Pr(\mathcal{X})$;  $\boldsymbol{X}_j \sim \Pr(\mathcal{X})$. 
\item The action taken was determined by a  policy that draws actions at random according to a (possibly unknown) probability distribution $p_0(\mathcal{A} | \boldsymbol{X}_j)$ on the action space $\mathcal{A}$.  (Note that the distribution of actions taken depends on the vector of features).
\item Only the reward of the action actually performed is recorded into the dataset, i.e., $R_j^{\text{obs}} \equiv R_j(A_j)$. 
\item For every action $a$, either taken or not taken, the reward $R_j(a) \sim \Phi_a(\cdot | \boldsymbol{X}_j)$ that would have been realized had $a$ actually been taken is generated by a random draw from an unknown family  $\{ \Phi_a( \cdot | \boldsymbol{x})\}_{ \boldsymbol{x} \in \mathcal{X}, a \in \mathcal{A}}$ of reward distributions with support $\left[0,1\right]$.
\end{enumerate}
The logging policy corresponds to the choices made by the existing decision-making procedure and so will typically create a biased distribution on the space of feature-action pairs. 

We make two natural assumptions about the rewards and the logging policy; taken together they enable us to generate unbiased estimates of the variables of the interest. The first assumption guarantees that there is enough information in the data-generating process so that counterfactual information can be inferred from what is actually observed. 

\begin{assumption}\label{ass:common}(Common support) $p_0(a | \boldsymbol{x}) > 0$ for all action-feature pairs $(a, \boldsymbol{x})$. 
\end{assumption}

The second assumption is that the logging policy depends only on the observed features -- and not on the observed rewards.  

\begin{assumption} \label{ass:unconfoundness}(Unconfoundness) For each feature vector $\boldsymbol{X}$, the rewards of actions $\{ R(a) \}_{a \in \mathcal{A}}$ are statistically independent of the action actually taken; $\{R(a)\} \independent A \big| \boldsymbol{X}$.  
\end{assumption}

These assumptions are universal in the counterfactual inference literature -- see ~\cite{johansson2016learning,athey2015machine} for instance -- although they can be criticized on the grounds that their validity cannot be determined on the basis of what is actually observed.

\section{The Algorithm}\label{sec:algorithm}

It seems useful to begin with a brief overview; more details and formalities follow below.  Our algorithm consists of a training phase and an execution phase; the training phase consists of three steps.
\begin{enumerate}[A.]
\item In the first step of the training phase, the algorithm either inputs the true propensity scores (if they are known) or uses the logged data to estimate propensity scores (when the true propensity scores are not known); this (partly) corrects the bias in the logged data.  
\item In the second step of the training phase, the algorithm uses the known or estimated propensity scores to compute, for each action and each feature, an estimate of relevance for that  feature with respect to that action. The algorithm then retains the more relevant features -- those for which the estimate is above a  threshold -- and discards the less relevant features -- those for which the estimate is below the threshold. (For reasons that will be discussed below, the threshold used depends on both the action and the feature type.)
\item In the third step of the training phase, the algorithm uses the known or estimated propensity scores and the features identified as relevant, and trains a feedforward neural network model to learn a non-linear stochastic policy that minimizes the "corrected" cross entropy loss. 
\end{enumerate}

In the execution phase, the algorithm is presented with a new instance and uses the policy derived in the training phase to recommend an action for this new instance on the basis of the relevant features of that instance.

Not surprisingly, the setting in which the  propensity scores are  known is  simpler than the setting in which the  propensity scores must be estimated.  In the latter case, in addition to the complication of the estimation itself, we shall need to be careful about estimated propensity scores that are ``too small'' -- this will require a correction -- and our error estimates will be less good.  Because clarity of exposition seems more importance than compactness, we therefore present first the algorithm for the case in which true propensity scores are known and then circle back to present the necessary modifications for the case in which true propensity scores are not known but must be estimated.

\subsection{True Propensities}
We begin with the setting in which propensities of the  randomized algorithm are actually tracked and available in the dataset. This is often the case in the advertising context, for example. In this case, for each $j$, set $p_{0,j} = p_0(A_j|X_j)$, and  write ${\boldsymbol{P}}_0 = [{p}_{0,j}]_{j=1}^n$; this is the vector of {\em true propensities}.  

\subsection{Relevance} 
It might seem natural to define the set $\mathcal{S}$ of feature types to be {\em irrelevant} (for a particular action) if the distribution of rewards (for that action) is independent of the features in $\mathcal{S}$, and to define the set $\mathcal{S}$ to be {\em relevant} otherwise. In theoretical terms, this definition has much to recommend it. In operational terms, however, this definition is not of much use.  That is because finding irrelevant sets of feature types would require many observations (to determine the entire distribution of rewards) and intractable calculations (to examine all sets of feature types). Moreover, this notion of irrelevance will often be too strong because our interest will often be only in maximizing expected rewards (or more generally some statistical function of rewards), as it would be in the medical context if the reward is five-year survival rate, or in the advertising or financial settings, if the reward is expected revenue or profit and the advertiser or firm is risk-neutral.

Given these objections, we take an alternative approach. We define a measure of how relevant a particular feature type is for the expected reward of a particular action, learn/estimate this measure from observed data, retain features for which this measure is above some endogenously derived threshold (the most relevant features) and discard other features (the least relevant features).  Of course, this approach has drawbacks.  Most obviously, it might happen that two feature types are individually not very relevant but are jointly quite relevant. (We leave this issue for future work.) However, as we show empirically, this approach has the virtue that it works: the algorithm we develop on the basis of this approach is demonstrably superior to existing algorithms.  

 \subsubsection{True Relevance}
 To begin formalizing our measure of relevance, fix an action $a$, a feature vector $x$ and a feature type $i$. Define  expected rewards and marginal expected rewards as follows:
 \begin{eqnarray}
 \bar{r}(a, \boldsymbol{x}) &=& \mathbb{E}\left[ R(a) | \boldsymbol{X} = \boldsymbol{x}\right] \nonumber \\
 \bar{r}(a, \boldsymbol{x}_i) &=&\mathbb{E}_{\boldsymbol{X}_{-i}} [\bar{r}(a, \boldsymbol{X}) \bigg| \boldsymbol{X}_{i} = \boldsymbol{x}_{i}] \nonumber \\
 \bar{r}(a) &=& \mathbb{E}_{\boldsymbol{X}} \left[ \bar{r}(a, \boldsymbol{X})\right]  \label{eqn:expectedreward}
 \end{eqnarray}
 We define the {\em true relevance of feature type $i$  for action $a$} by

 \begin{align}
 g(a, i) = \mathbb{E}\left[ \ell\left(\bar{r}(a, X_i) - \bar{r}(a)\right) \right], \label{eqn:rel_metric}
 \end{align}
where the expectation is taken with respect to the arrival probability distribution of $X_i$ and $\ell(\cdot)$ denotes the loss metric. (Keep in mind that the true arrival probability distribution of $X_j$ is unknown and must be estimated from the data.) Our results hold for an arbitrary loss function, assuming only that it is strictly monotonic and Lipschitz; i.e. there is a constant $B$ such that $\left| \ell(r) - \ell(r')\right| \leq B |r - r'|$. These conditions are satisfied by a large class of loss functions including $l_1$ and $l_2$ losses. The relevance measure $g$ expresses the weighted difference between the expected reward {\em of a given action} conditioned on the feature type $i$ and the unconditioned expected reward; $g(a,i) = 0$ exactly when   feature type $i$ does not affect the expected reward of action $a$.\footnote{Other measures of relevance have been used in the feature selection literature (e.g., especially Pearson correlation~\cite{hall1999correlation} and mutual information~\cite{yu2003feature}) -- but not for relevance of actions.} 

We refer to $g$ as  {\em true} relevance because it is computed using the {\em true} arrival distribution -- but the true arrival distribution is unknown. Hence, even when the true propensities are known, relevance must be {\em estimated} from observed data. This is the next task. 

\subsubsection{Estimated Relevance}
We now derive {\em estimates} of relevance based on observed data (continuing to assume  that true propensities are known). To do so, we first need to estimate $\bar{r}(a)$ and $\bar{r}(a, x_i)$ for $x_i \in \mathcal{X}_i$, $i \in \mathcal{F}$ and $a \in \mathcal{A}$ from available observational data. An obvious way to do this is through classical supervised learning based estimators; most obviously, the sample mean estimators for $\bar{r}(a)$ and $\bar{r}(a, x_i)$.  However using straightforward sample mean estimation would be wrong because the logging policy introduces a bias into observations. Following the idea of Inverse Propensity Scores ~\cite{rosenbaum1983central}, we correct this bias by using Importance Sampling.  
 
Write $N(a)$, $N(x_i)$, $N(a, x_i)$ for the number of observations (in the given data set) with action $a$, with feature $x_i$, and with the pair consisting of action $a$ and feature $x_i$, respectively. We can rewrite our previous definitions as:
\begin{eqnarray}
\bar{r}(a, x_i) &=& \mathbb{E}_{(\boldsymbol{X}, A, R^{\text{obs}}) \sim p_0} \left[ \frac{\mathbb{I}(A = a)R^{\text{obs}} }{p_0(A | \boldsymbol{X})} \bigg| X_i = x_i \right] \notag \\ 
 \bar{r}(a) &=& \mathbb{E}_{(\boldsymbol{X}, A, R^{\text{obs}}) \sim p_0} \left[ \frac{\mathbb{I}(A = a) R^{\text{obs}}}{p_0(A | \boldsymbol{X})}  \right] \label{eqn:IS_2}
\end{eqnarray}
where $\mathbb{I}(\cdot)$ is the indicator function. (Note that we are taking expectations with respect to the true propensities.)  

Let $\mathcal{J}(x_i)$ denote the time indices in which feature type-$i$ is $x_i$, i.e., $\mathcal{J}(x_i) = \{ j \subseteq \{1,2, \ldots,n\} : X_{i,j} = x_i \}$. The Importance Sampling approach provides unbiased estimates of $\bar{r}(a)$ and $\bar{r}(a, x_i)$ as
\begin{eqnarray}
\widehat{R}(a, x_i ; \boldsymbol{P}_0) &=& \frac{1}{N(x_i)}\sum_{j \in \mathcal{J}(x_i)} \frac{ \mathbb{I}(A_j = a) R^{\text{obs}}_j}{p_{0,j}}, \notag \\ 
\widehat{R}(a ; \boldsymbol{P}_0) &=& \frac{1}{n}\sum_{j=1}^{n} \frac{\mathbb{I}(A_j = a) R^{\text{obs}}_j }{p_{0,j}} ,\label{eqn:propensity_1}
\end{eqnarray}
(We include the propensities $\boldsymbol{P}_0$ in the notation as a reminder that these estimators are using the {\em true} propensity scores.)

We  now define the {\em estimated relevance of feature type $i$  for action $a$} as 
\begin{align}
\widehat{G}(a, i; \boldsymbol{P}_0) = \frac{1}{n}\sum_{x_i \in \mathcal{X}_i} N(x_i)  \ell \left(\widehat{R}(a, x_i ; \boldsymbol{P}_0) - \widehat{R}(a ; \boldsymbol{P}_0) \right). \label{eqn:rel_metric}
\end{align} 
(Note that we have abused terminology/notation by suppressing reference to the particular sample that was observed.)

\subsubsection{Thresholds}

By definition, $\widehat{G}$ is an estimate of relevance so the obvious way to select relevant features is to set a threshold $\tau$, identify a feature $i$ as relevant for action $a$ exactly when $\widehat{G}(a, i; \boldsymbol{P}_0) > \tau$, retain the features that are relevant according to this criterion and discard other features.  

However, this approach is a bit too naive for (at least) two reasons. The first is that our empirical estimate of relevance $\widehat{G}$ may in fact be far from the true relevance $g$. The second is that some features may be highly (positively or negatively) correlated with the remaining features, and hence convey less information. To deal with these objections, we construct thresholds $\tau(a,i)$ as a weighted sum of an empirical estimate of the error in using $\widehat{G}$ instead of $g$ and the (average absolute) correlation of feature type $i$ with other feature types.

To define the first term we need an empirical (data-dependent bound) on $|\widehat{G} - g|$.  To derive such a bound we use the empirical Bernstein inequality~\cite{maurer2009empirical,audibert2009exploration}. (We emphasize that our bound depends on the \textit{empirical variance} of the estimates.)  
To simplify notation,  define random variables $U(a; \boldsymbol{P}_0) \equiv \frac{\mathbb{I}(A = a) R^{\text{obs}}}{p_0(A | \boldsymbol{X})}$ and $U_j(a; \boldsymbol{P}_0) \equiv \frac{\mathbb{I}(A_j = a) R_j}{p_{0,j}}$.  The sample means and variances are: 
\begin{eqnarray*}
\mathbb{E}_{(\boldsymbol{X}, A, R^{\text{obs}}) \sim p_0}[ U(a; \boldsymbol{P}_0) ] &=& \bar{r}(a), \notag \\
\mathbb{E}_{(\boldsymbol{X}, A, R^{\text{obs}}) \sim p_0}[ U(a; \boldsymbol{P}_0) \big| X_i = x_i ] &=& \bar{r}(a, x_i)
 \notag \\
 \widehat{U}(a; \boldsymbol{P}_0) &=& \widehat{R}(a ; \boldsymbol{P}_0) \\
     &=& \frac{1}{n}\sum_{j=1}^n U_j(a; \boldsymbol{P}_0), \notag \\ 
 \widehat{U}(a, x_i; \boldsymbol{P}_0) &=& \widehat{R}(a, x_i ; \boldsymbol{P}_0) \\
 &=& \frac{1}{N(x_i)}\sum_{j \in \mathcal{J}(x_i)} U_j(a; \boldsymbol{P}_0), \notag \\
 V_n(a ; \boldsymbol{P}_0) &=& \frac{1}{n-1}\sum_{j=1}^n \left(U_j(a ; \boldsymbol{P}_0) - \widehat{U}(a; \boldsymbol{P}_0)\right)^2, \notag \\ 
 V_n(a, x_i ; \boldsymbol{P}_0) &=& \frac{1}{N(x_i)-1}\sum_{j \in \mathcal{J}(x_i)} \left(U_j(a; \boldsymbol{P}_0) - \widehat{U}(a, x_i ; \boldsymbol{P}_0)\right)^2. \notag 
\end{eqnarray*}
The weighted average sample variance is: 
\begin{align}
\bar{V}_n(a, i ; \boldsymbol{P}_0) = \sum_{x_i \in \mathcal{X}_i} \frac{N(x_i) V_n(a, x_i ; \boldsymbol{P}_0)}{n}
\end{align}
Our empirical (data-dependent) bound is given in Theorem 1.

\begin{theorem} \label{thm:gen_bound} For every $n >0$, every $\delta \in \left[0,\frac{1}{3}\right]$, and every pair, $(a,i) \in \left(\mathcal{A}, \mathcal{D} \right)$,  with probability at least $1 - 3 \delta$ we have: 
\begin{eqnarray*}
|\widehat{G}(a,i ; \boldsymbol{P}_0) - g(a,i)| &\leq& B \Bigg( \ \sqrt{\frac{2  b_i   \ln (3/\delta) \bar{V}_n(a,i; \boldsymbol{P}_0)}{n}} \notag \\
&& \ \ + \ \sqrt{\frac{2  \ln (3/\delta) V_n(a; \boldsymbol{P}_0) }{n}} \notag \\
&& \ \ \ \ \ \ +   \  \frac{M \left(b_i + 1\right) \ln 3/\delta}{n} \ \Bigg)\notag \\ 
&& \ \ \ \ \ \ \ \ \ \ +\ \sqrt{\frac{2 \left( \ln 1/\delta + b_i \ln 2\right)}{n}}, \notag
\end{eqnarray*}
where $M = \max_{a \in \mathcal{A}} \max_{\boldsymbol{x} \in \mathcal{X}} 1/p_0(a | \boldsymbol{x})$. 
\end{theorem}

The error bound given by Theorem \ref{thm:gen_bound} consists of four terms: The first term arises from estimation error of $\widehat{R}(a, x_i)$. The second term arises from estimation error of $\widehat{R}(a)$. The third term arises from estimation error of feature arrival probabilities. The fourth term arises from  randomness of the  logging policy. 

Now write $\rho_{i,j}$ for the Pearson correlation coefficient between two feature types $i$ and $j$.  (Recall that $\rho_{i,j} = +1$ if $i, j$ are perfectly positively correlated, $\rho_{i,j} = -1$ if $i, j$ are perfectly negatively correlated, and $\rho_{i,j} = 0$ if $i, j$ are uncorrelated.)  Then the average absolute correlation of feature type $i$ with other features is
$$
\bigg(\frac{1}{d-1}\bigg) \bigg(\sum_{j \in \mathcal{F} \setminus \{i\}} \left|\rho_{i,j}\right| \bigg)
$$

We now define the thresholds as 
$$
\tau(a,i) = \lambda_1  \sqrt{\frac{  b_i \bar{V}_n(a,i; \boldsymbol{P}_0) }{n}} \ + \ \lambda_2 \bigg(\frac{1}{d-1}\bigg) \bigg(\sum_{j \in \mathcal{F} \setminus \{i\}} \left|\rho_{i,j}\right| \bigg)
$$
where $\lambda_1, \lambda_2$ are weights (hyper-parameters) to be chosen. Notice that the first term is the dominant term in the error bound given in Theorem 1, and is used to set a higher bar for the feature types that are creating the logging policy bias. The statistical distributions of those features within the the action population and the whole population will be different. By setting the threshold as above, we trade-off between three objective: (1) selecting the features that are relevant for the rewards of the actions, (2) eliminating the features which create the logging policy bias, (3) minimizing the redundancy in the feature space. 

\subsubsection{Relevant Feature Types}

Finally, we identify the set of feature types that are relevant for an action $a$ as
\begin{align}
\widehat{\mathcal{R}}(a) = \left\{ i \ : \  \widehat{G}(a,i; {\mathbf P}_0) > \tau(a,i) \right\} \label{eqn:rel_opt}
\end{align}
Set $\widehat{\boldsymbol{\mathcal{R}}} = \left[\widehat{\mathcal{R}}(a) \right]_{a \in \mathcal{A}}$. Let $\boldsymbol{f}_a$ denote a $d$ dimensional vector whose $j^{th}$ element is $1$ if $j$ is contained in the set $\mathcal{R}(a)$ and $0$ otherwise.

\begin{figure}
\centering
    \includegraphics[width=1\textwidth]{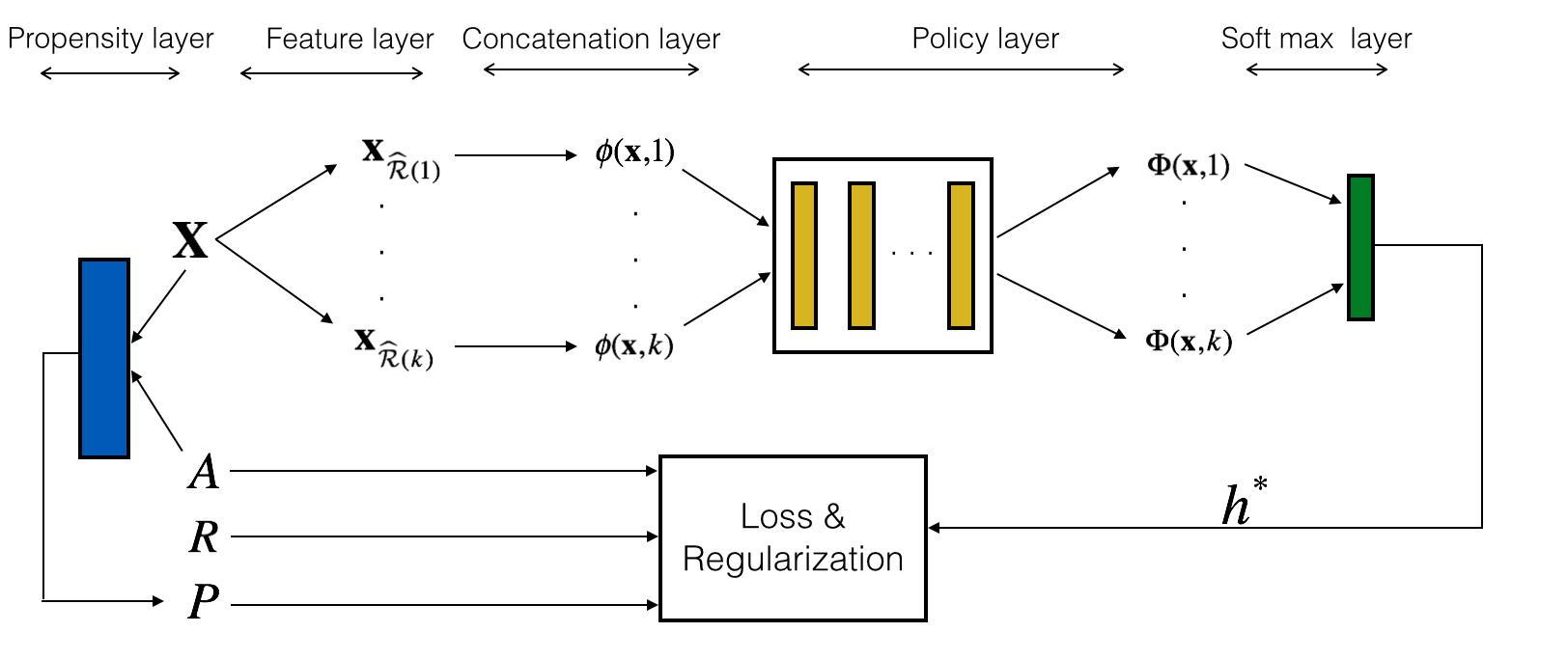}
    \caption{Neural network architecture}
\end{figure}

\subsection{Policy Optimization} 
We now build on the identified family of relevant features to construct a policy. By definition, a (stochastic) policy is a map $h : \mathcal{X} \rightarrow \triangle(A)$ which assigns to each vector of features a probability distribution $h(\cdot | \boldsymbol{x})$ over actions. 

A familiar approach to the construction of stochastic policies is to use the POEM algorithm~\cite{swaminathan2015batch}. POEM considers only linear stochastic policies; among these, POEM learns one that minimizes risk, adjusted by a variance term. Our approach is substantially more general because we consider arbitrary non-linear stochastic policies. \rev{We use a novel approach that uses a feedforward neural network to find a non-linear policy that minimizes the loss, adjusted by a regularization term. Note that we allow for very general loss and regularization terms so that our approach includes many policy optimizers.  If we restricted to a neural network with no hidden layers and a specific regularization term, we would recover POEM. }

\rev{ We propose a feedforward neural network for learning a policy $h^{*}(\cdot | \boldsymbol{x})$; the architecture of our neural network is depicted in Fig. 1. Our feedforward neural network consists of policy layers ($L_p$ hidden layers with  $h_p^{(l)}$  units in the $l^{\text{th}}$ layer) that use the output of the concatenation layer to generate a policy vector $\Phi(\boldsymbol{x}, a)$, and a softmax layer that turns the policy vector into a stochastic policy.

For each action $a$, the concatenation layer takes the feature vector $\boldsymbol{x}$ as an input and generates a action-specific representations $\phi(\boldsymbol{x}, a)$ according to:  
 \begin{eqnarray*}
 \boldsymbol{x}_{\widehat{\mathcal{R}}(a)} &=& \boldsymbol{x} \odot \boldsymbol{f}_a \notag \\ 
 \phi(\boldsymbol{x}, a) & = & [\boldsymbol{x}_{\widehat{\mathcal{R}}(\tilde{a})} \mathbb{I}(\tilde{a} = a)]_{\tilde{a} \in \mathcal{A}}
 \end{eqnarray*}
Note that our action-specific representation $\phi(\boldsymbol{x}, a)$ is a $d \times k$ dimensional vector where only the parts corresponding to action $a$ is non-zero and equals to $\boldsymbol{x}_{\widehat{\mathcal{R}}(\tilde{a})}$.} For each action $a$, the policy layers uses the action-specific representation $\phi(\boldsymbol{x}, a)$ generated by the concenation layers and generates the output vector $\Phi(\boldsymbol{x},a)$ according to: 
$$
\Phi(\boldsymbol{x}, a)  =  \rho\left( \ldots \rho\left( \boldsymbol{W}_1^{(p)} \phi(\boldsymbol{x}, a)  + \boldsymbol{b}_1^{(p)} \right) \ldots + \boldsymbol{b}_{L_p}^{(p)} \right)
$$
where $\boldsymbol{W}_l^{(p)}$ and $\boldsymbol{b}_l^{(p)}$ are the weights and bias vectors of the $l^{\text{th}}$ layer accordingly. The outputs of the policy layers are used to generate a policy by a  softmax layer: 
$$
h(a | \boldsymbol{x}) = \frac{\exp( \boldsymbol{w}^T \Phi(\boldsymbol{x}, a) )}{\sum_{a' \in \mathcal{A}} \exp( \boldsymbol{w}^T \Phi(\boldsymbol{x}, a') )}.  \notag \\ 
$$

\rev{Then, we choose the parameters of the policy to minimize an objective of the following form: $\text{Loss}(h^{*}; \mathcal{D}) + \lambda_3 \mathcal{R}(h^{*}; \mathcal{D})$; where $\text{Loss}(h^{*}; \mathcal{D})$ is the loss term, $\mathcal{R}(h^{*}; \mathcal{D})$ is a regularization term and  $\lambda_3 >0$ represents the trade-off between loss and regularization. The loss function can be either the negative IPS estimate or the corrected cross entropy loss introduced in the next section. Depending on the choice of the loss function and regularizer, our policy optimizer can include a wide-range of objectives including the POEM objective ~\cite{swaminathan2015batch}. 

In the next subsection, we propose a new objective, which we refer to as the Policy Neural Network (PONN) objective. 
}

\subsection{Policy Neural Network (PONN) objective}

Our PONN objective is motivated by the cross-entropy loss used in the standard multi-class classification setting. In the usual classification setting, usual loss function used to train the neural network is the standard cross entropy: 
$$
\widehat{\text{Loss}}_{c}(h) = \frac{1}{n} \sum_{j=1}^n \sum_{a \in \mathcal{A}} -R_j(a) \log h(a | \boldsymbol{X}_j). 
$$
However, this loss function is not applicable in our setting, for two reasons. The first is that only the rewards of the action taken by the logging policy are recorded in the dataset, not the counterfactuals. The second is that we need to correct the bias in the dataset by weighting the instances by their inverse propensities. Hence, we use the following modified cross entropy loss function: 
\begin{eqnarray}
\widehat{\text{Loss}}_{b}(h; \boldsymbol{P}_0) &=& \frac{1}{n} \sum_{j=1}^n \sum_{a \in \mathcal{A}} \frac{- R_j(a) \log h(a | \boldsymbol{X}_j) \mathbb{I} (A_j =a)}{p_{0,j}} \notag \\ 
&=& \frac{1}{n} \sum_{j=1}^n  \frac{- R^{\text{obs}}_j \log h(A_j | \boldsymbol{X}_j)}{p_{0,j}}. \label{eqn:cross_entropy}
\end{eqnarray}
Note that this loss function is an unbiased estimate of the expected cross entropy loss, that is $\mathbb{E}_{(\boldsymbol{X}, A, R) \sim p_0}\left[ \widehat{\text{Loss}}_{b}(h^{*}; \boldsymbol{P}_0) \right] = \mathbb{E} \left[ \widehat{\text{Loss}}_{c}(h^{*}) \right]$. We train our neural network to minimize the regularized loss by Adam optimizer: 
$$
h^{*}= \argmin_{h \in \mathcal{H}}\; \widehat{\text{Loss}}_{b}(h; \widehat{\boldsymbol{P}}_0)+ \lambda_3 \mathcal{R}(h), \label{eqn:CRM}
$$
where $\mathcal{R}(h)$ is the regularization term to avoid overfitting and $\lambda_3$ is the hyperparameter to trade-off between the loss and regularization.

\subsection{Unknown Propensities}
As we have noted, in most settings the logging policy is unknown and hence the actual propensities are also unknown so we  must {\em estimate} propensities from the dataset and use the {\em estimated} propensities to correct the bias. In general, this can be accomplished by any supervised learning technique.  

For our purposes we estimate propensities by fitting the multinomial logistic regression model:
\begin{align}
\ln( \Pr\left( A = a \right)) = \boldsymbol{\beta}_{0,a}^T \boldsymbol{X} - \ln Z \label{eqn:estimate_propensities}
\end{align}
where $Z = \sum_{a \in \mathcal{A}} \exp\left( \boldsymbol{\beta}_{0,a}^T \boldsymbol{X} \right)$. The estimated propensities are 
$$
\widehat{p}_{0,j} \equiv \frac{\exp(\boldsymbol{\beta}_{0,A_j}^T \boldsymbol{X}_j)}{Z_j}
$$ 
where we have written $Z_j = \sum_{a \in \mathcal{A}} \exp(\boldsymbol{\beta}_{0,a}^T \boldsymbol{X}_j)$.  Write 
$\widehat{\boldsymbol{P}}_0 = [\widehat{p}_{0,j}]_{j=1}^n$ for the vector of  estimated propensities 

In principle, we could use these estimated propensities in place of known propensities and proceed exactly as we have done above.  However, there are two problems with doing this.  The first is that if the estimated propensities are very small (which might happen because the data was not completely representative of the true propensities), the variance of the estimate $\widehat{G}$  will be too large.  The second is that the thresholds we have constructed when propensities are known may no longer be appropriate when propensities must be estimated.

To avoid the first problem, we follow ~\cite{ionides2008truncated} and modify the estimated rewards by truncating the importance sampling weights. This leads to ``truncated'' estimated rewards as follows:
\begin{eqnarray*}
 \widehat{R}_m(a, x_i; \widehat{\boldsymbol{P}}_0) &=& \frac{1}{N(x_i)}\sum_{j \in \mathcal{J}(x_i)} \min\left(\frac{\mathbb{I}(A_j = a)}{\widehat{p}_{0,j}}, m \right) R_j^{\text{obs}}, \notag \\ 
 \widehat{R}_m(a; \widehat{\boldsymbol{P}}_0) &=& \frac{1}{n}\sum_{j =1}^n \min\left(\frac{\mathbb{I}(A_j = a)}{\widehat{p}_{0,j}}, m \right) R_j^{\text{obs}}. \notag
\end{eqnarray*}
Given these ``truncated'' estimated rewards, we define a ``truncated''  estimator of relevance by
\begin{align}
\widehat{G}_m(a, i ; \widehat{\boldsymbol{P}}_0) = \sum_{x_i \in \mathcal{X}_i} \frac{N(x_i)}{n} l \left( \widehat{R}_m(a, x_i ; \widehat{\boldsymbol{P}}_0) - \widehat{R}_m(a; \widehat{\boldsymbol{P}}_0) \right) \notag 
\end{align}

From this point on, we proceed exactly as before, using the ``truncated'' estimator $\widehat{G}_m$ instead of $\widehat{G}$.

Note that $\widehat{R}_m(a, x_i; \widehat{\boldsymbol{P}}_0)$ and $\widehat{R}_m(a; \widehat{\boldsymbol{P}}_0)$ are not unbiased estimators of $\bar{r}(a, x_i)$ and $\bar{r}(a)$. The bias is due to using estimated truncated propensity scores which may deviate from true propensities. Let $\operatorname{bias}(\widehat{R}_m(a ; \widehat{\boldsymbol{P}}_0))$ denote the bias of $\widehat{R}_m(a ; \widehat{\boldsymbol{P}}_0)$, which is given by
$$
\operatorname{bias}(\widehat{R}_m(a ;  \widehat{\boldsymbol{P}}_0)) = \bar{r}(a) - \mathbb{E}\left[\widehat{R}_m(a ; \widehat{\boldsymbol{P}}_0)\right].
$$

In the Appendices, we show the effect of this bias on the learning process.

\begin{algorithm}[t]
\caption{Training Phase of the Algorithm PONN-B}
	\label{alg:LADR}
	\normalsize
	\begin{algorithmic}[1]
		\STATE {\bfseries Input:} $\lambda_1, \lambda_2, \lambda_3$, $L_r$, $L_p$, $h_i^{r}$, $h_j^{a}$
		\STATEx {\bfseries Step A: Estimate propensities using a logistic regression}
		\STATE Compute $\boldsymbol{\beta}_{0,a}$ for each $a$ by training Logistic regression model from~(\ref{eqn:estimate_propensities}).
		\STATE Set $\widehat{p}_{0,j} = \exp(\boldsymbol{\beta}_{0,A_j}^T \boldsymbol{X}_j)/Z_j$ with $Z_j = \sum_{a \in \mathcal{A}} \exp(\boldsymbol{\beta}_{0,a}^T \boldsymbol{X}_j)$.
		\STATEx{\bfseries Step B: Identify the relevant features}
		\STATE Compute $\widehat{R}(a, x_i; \widehat{\boldsymbol{P}}_0)$, $\widehat{R}(a; \widehat{\boldsymbol{P}}_0)$, $\bar{V}_n(a,i; \widehat{\boldsymbol{P}}_0)$, $\rho_{i,l}$ for each $a$, $x_i$, $i$, $l$. 
		\STATE Compute $\widehat{G}(a, i; \widehat{\boldsymbol{P}}_0)$ for each action-feature type pair.
		\STATE Solve $\widehat{\mathcal{R}}(a)$ from~(\ref{eqn:rel_opt}). 
		\STATEx {\bfseries Step C: Policy Optimization}
		\WHILE{$\text{until convergence}$ }
		\STATE {$\left( \boldsymbol{w}, \boldsymbol{W}_p^{(l)}\right) \leftarrow \operatorname{Adam}\left(\mathcal{D}^{(n)}, \boldsymbol{w}, \boldsymbol{W}_p^{(l)}\right)$
		\ENDWHILE}
		\STATEx{\bfseries Output of Training Phase:} Policy $h^{*}$,  Features $\widehat{\boldsymbol{\mathcal{R}}}$ 		
	\end{algorithmic}
\end{algorithm}

\begin{algorithm}[t]
\caption{Execution Phase of the Algorithm PONN-B}
	\label{alg:LADR}
	\normalsize
	\begin{algorithmic}[1]
	\STATE{\bfseries Input: }Instance with feature $\boldsymbol{X}$ \\
	\color{red}{Set  $\hat{a}(\boldsymbol{X}) = \arg\max_{a \in \mathcal{A}} h^{*}(a | \boldsymbol{X})$}
	\STATEx {\bfseries Output of Execution phase:} Recommended action $\hat{a}(\boldsymbol{X})$
	\end{algorithmic}
\end{algorithm}

\section{Pseudo-code for the Algorithm PONN-B} \label{sec:algorithm}  
Below, we provide the pseudo-code for our algorithm which we call PONN-B (because it uses the PONN objective and Step B) exactly as discussed above.  The first three steps constitute the offline training phase; the fourth step is the online execution phase. Within the training phase the steps are: Step A: Input propensities (if they are known) or  estimate them using a logistic regression (if they are not known).  Step B: Construct estimates of relevance (truncated if propensities are estimated), construct thresholds (using given hyper-parameters) and identify the relevant features as those for which the estimated relevance is above the constructed thresholds. Step C: Use the Adam optimizer to train neural network parameters. In the execution phase: Input the features of the new instance, apply the optimal policy  to find a probability distribution over actions, and draw a random sample action from this distribution.

\section{Extension: Relevant Feature Selection with Fine Gradations }~\label{sec:ext}
Our algorithm might be inefficient when there are many features of a particular type -- in particular, if one or more feature types are continuous.  In that setting, we can modify our algorithm  to create  bins that consist of \textit{similar} feature values and treat all the values in a single bin identically. In order to conveniently formalize this problem, we assume that the feature space is actually continuous; for simplicity we  assume each feature type is $\mathcal{X}_i = \left[0,1\right]$ (or a bounded subset). In this case, we can partition the feature space into subintervals  (bins), view features in each bin as identical, and apply our algorithm to the finite set of bins.\footnote{The binning procedure loses the ordering in the interval $\left[0,1\right]$.  If this ordering is in fact relevant to the feature, then the binning procedure loses some information that a different procedure might preserve.  We leave this for future work.} To offer a theoretical justification for this procedure, we assume that similar features yield similar expected rewards.  We formalize this as a Lipschitz condition.

\begin{assumption} \label{ass:sim} There exists $L > 0$ such that for all $a \in \mathcal{A}$, all $i \in \mathcal{F}$ and all $x_i \in \mathcal{X}_i$, we have $|\bar{r}(a, x_i) - \bar{r}(a, \tilde{x}_i)| \leq L |x_i - \tilde{x}_i|$. 
\end{assumption}
(In the Multi-Armed Bandit literature~\cite{slivkins2014contextual,tekin2014discovering} this assumption is commonly made and sometimes referred to as {\em similarity}.)

For convenience, we partition each feature type $X_i$ into $s$ equal subintervals (bins) of length $1/s$.  If $s$ is small, the number of bins is small so, given a finite data set, the number of instances that lie in each bin is relatively large; this is useful for estimation.  However, when $s$ is small the size $1/s$ of each bin is relatively large so the (true) variation of expected rewards in each bin is relatively large.   Because we are free to choose the parameter $s$, we can balance the trade-off implicit between choosing few large bins or choosing many small bins; a useful trade-off is achieved by taking $s = \left \lceil{n^{1/3}}\right \rceil$.

So begin by fixing $s = \left \lceil{n^{1/3}}\right \rceil$ and partition each $\mathcal{X}_i =\left[0,1\right]$ into  $s$ intervals of length $1/s$. Write 
$\mathcal{C}_{i}$ for the sets in the partition of $X_i$ and write $c_i$ for a typical element of $\mathcal{C}_{i}$ .  For each sample $j$, let $c_{i,j}$ denote the set in which the feature $x_{i,j}$ belongs. Let $\mathcal{J}(c_i)$  be the set of indices for which $x_{i,j} \in c_i$; $\mathcal{J}(c_i) = \{ j \in \{1,2,\ldots,n\}: X_{i,j} \in c_i \}$. We define truncated IPS estimate as 
\begin{eqnarray*}
 \bar{r}_m(a, c_i; \widehat{\boldsymbol{P}}_0) &=& \mathbb{E}\left[ U(a; \widehat{\boldsymbol{P}}_0) | X_i \in c_i \right] \notag \\ 
&=& \mathbb{E}\left[ \min\left( \frac{\mathbb{I}(A = a)}{\widehat{p}_0(A | \boldsymbol{X})}, m\right) R^{\text{obs}} \bigg| X_i \in c_i \right], \notag \\ 
 \widehat{R}_m(a, c_i ; \widehat{\boldsymbol{P}}_0) &=&\frac{1}{N(c_i)} \sum_{j \in \mathcal{J}(c_i)} \min\left( \frac{\mathbb{I}(A_j = a) }{\widehat{p}_{0,j}}, m \right) R_j^{\text{obs}}, \notag
\end{eqnarray*}
where $N(c_i) =|\mathcal{J}(c_i)|$. In this case, we define estimated information gain as 
\begin{align}
\widehat{G}_m(a, i) = \sum_{c_i \in \mathcal{C}_i} \frac{N(c_i)}{n} l \left( \widehat{R}_m(a, c_i ; \widehat{\boldsymbol{P}}_0) - \widehat{R}_m(a ; \widehat{\boldsymbol{P}}_0) \right). \notag 
\end{align} 
We define the following sample mean and variance : 
\begin{align}
& \widehat{U}(a, c_i ; \widehat{\boldsymbol{P}}_0) = \widehat{R}_m(a, c_i ; \widehat{\boldsymbol{P}}_0) = \frac{1}{N(x_i)}\sum_{j \in \mathcal{J}(c_i)} U_j(a; \widehat{\boldsymbol{P}}_0), \notag \\ 
& V_n(a, c_i ; \widehat{\boldsymbol{P}}_0) = \frac{1}{n-1}\sum_{j \in \mathcal{J}(c_i)} (U_j(a,c_i; \widehat{\boldsymbol{P}}_0)- \widehat{U}(a, c_i ;\widehat{\boldsymbol{P}}_0))^2. \notag 
\end{align}
Let $\bar{V}_n(a, i ; \widehat{\boldsymbol{P}}_0) = \sum_{c_i \in \mathcal{C}_i} \frac{N(c_i) V_n(a, c_i ; \widehat{\boldsymbol{P}}_0) }{n}$ denote the weighted average sample variance. 

Given these definitions, we establish a data-dependent bound analogous to Theorem 1.
\begin{theorem} \label{thm:gen_bound_cont} For every $n \geq 1$ and $\delta \in \left[0,\frac{1}{3}\right]$,  if $s = \left \lceil{n^{1/3}}\right \rceil$, then  with probability at least $1 - 3 \delta$ we have, for all pairs $(a,i) \in \left(\mathcal{A}, \mathcal{D} \right)$, 
\begin{eqnarray*}
 |\widehat{G}_m(a,i ; \widehat{\boldsymbol{P}}_0) - g(a,i)|  
& \leq& B \Bigg( \frac{\sqrt{4 \ln 3/\delta}}{n^{1/3}} \left( \sqrt{\bar{V}_n(a,i ; \widehat{\boldsymbol{P}}_0)} + \sqrt{V_n(a ;\widehat{\boldsymbol{P}}_0)}\right) + \frac{L}{n^{1/3}} \notag \\ 
&& \ \ \ \  + \left|\operatorname{bias}(\widehat{R}_m(a ; \widehat{\boldsymbol{P}_0}))\right| + \mathbb{E} \left|\operatorname{bias}(\widehat{R}_m(a, X_i ; \widehat{\boldsymbol{P}_0}))\right|  \Bigg) \notag \\
&& \ \ \ \ \ + \ \frac{4m B \ln 3/\delta +  \sqrt{2 \ln 1/\delta + \ln 2}}{n^{2/3}}. 
\end{eqnarray*} 
\end{theorem}
There are two main differences between Theorem 1 and Theorem 2. The first is that the estimation error  is decreasing as $n^{1/3}$ (Theorem 2) rather than as $n^{1/2}$ (Theorem 1). The second is that there is an additional error in Theorem 2 arising from the Lipschitz bound.

Theorem 2 suggests a different choice of thresholds, namely:
\begin{eqnarray*}
\tau(a,i) &=&\lambda_1 n^{-1/3}\sqrt{V_n(a,i; \widehat{\boldsymbol{P}}_0)} + \lambda_2\left(\frac{1}{d-1}\right) \left( \sum_{l \in \mathcal{F} \setminus \{i\}} \left|\rho_{i,l}\right|  \right). \notag
\end{eqnarray*}

With this change we proceed exactly as before. 

\section{Numerical Results}~\label{sec:num} 

Here we describe the performance of our algorithm on some real datasets.  Note that it is difficult (perhaps impossible) to validate and test the algorithm on the basis of actual logged CMAB data unless the counterfactual action rewards for each instance are available -- which would (almost) never  be the case. One way to validate and test our algorithm is to use a multi-class classification dataset, generate a biased CMAB dataset for training by ``forgetting'' (stripping out) the counterfactual information, apply the algorithm, and then test the predictions of the algorithm against the actual data~\cite{beygelzimer2009offset}. This is the route we follow in the first experiment below.  Another way to validate and test our algorithm is to use an alternative accepted procedure to infer counterfactuals and to test the prediction of our algorithm against this alternative accepted procedure. This is the route we follow in the second experiment below.

\begin{table*}[] 
\caption{Data Summary} 
\label{table:summary} 
\normalsize
\centering 
\begin{tabular}{|c|c|c|c|} 
\hline 
Dataset   & \# of Feature types (d) & \# of Labels (k) & \# of Instances (n) 
\\ 
\hline 
pendigits & 16                      & 10               & 7494                \\ \hline 
satimage  & 36                      & 6                & 4435                \\ \hline 
optdigits & 64                      & 10               & 3893                \\ \hline 
\end{tabular} 
\end{table*}

\subsection{Multi-class classification}
For this experiment we use existing multi-class classification datasets from the well-known UCI Machine Learning Repository. 

\begin{itemize}
\item In the {\bf Pendigits} and {\bf Optdigits} datasets, each instance is described by a collection of pixels extracted from the image of a handwritten digit 0-9; the objective is to identify the digit from the features.
\item In the {\bf Satimage} dataset, each instance is described by an array of features extracted from a satellite image of a plot of ground; the objective is to identify the true description of the plot (barren soil, grass, cotton crop, etc.) from the features.
\end{itemize}
These datasets have in common that that they have many instances, many feature types and many labels, so they are extremely useful for training and testing. 

In supervised learning systems, we assume that features and labels are generated by an i.i.d. process, i.e., $\left( \boldsymbol{X}, Y\right) \sim Z$ where $\boldsymbol{X} \in \mathcal{X}$ is the feature space and $Y \in \{1,2,\ldots, k \}$ is the label space. The supervised learning data with $n$-samples is denoted as $\mathcal{D}^n = \left(\boldsymbol{X}_j, Y_j\right)_{j=1}^n$. In our simulation setup, we treat each class as an action. We also included $16$ irrelevant features in addition to actual features in the dataset, drawn randomly from normal distribution. The reward of an action is given by $R_j(a) = \mathbb{I}(Y_j = a)$. A complete dataset then is $\mathcal{D}^n_{\text{com}} = \left(\boldsymbol{X}_j, R_j(1), \ldots, R_j(k)\right)$.  A summary of the data is given in Table \ref{table:summary}.
 
\subsection{Comparisons}
 We compare the performance of our algorithm (PONN-B) with 
 \begin{itemize}
\item \textbf{PONN} is PONN-B but  {\em without} Step B (feature selection). 
\item \textbf{POEM} is the standard POEM algorithm ~\cite{swaminathan2015batch}.
\item \rev{\textbf{POEM-B} applies Step B of our algorithm, followed by the POEM algorithm.}
\item \textbf{POEM-L1} is the POEM algorithm  with the addition of $L_1$ regularization. 
\item \textbf{Multilayer Perceptron with $L_1$ regularization (MLP-L1)} is the MLP algorithm on concatenated input $(\boldsymbol{X}, A)$ with $L_1$ regularization. 
\item \textbf{Logistic Regression with $L_1$ regularization (LR-L1)} is the separate LR algorithm on input $\boldsymbol{X}$ on each action $a$ with $L_1$ regularization. 
\item \textbf{Logging} is the logging policy performance. 
\end{itemize}
 (In all cases, the objective is optimized with the Adam Optimizer.)
 
\subsubsection{Simulation Setup} 
We generate artificially biased dataset by the following logistic model. We first draw weights for each label from an multivariate Gaussian distribution, that is $\theta_{0,y} \sim \mathcal{N}(0, \kappa I)$.  We then use the logistic model to generate an artificially biased logged off-policy dataset $\mathcal{D}^n = \left(\boldsymbol{X}_j, A_j, R_j^{\text{obs}} \right)_{j=1}^n$ by first drawing an action $A_j \sim p_0( \cdot | \boldsymbol{X}_j)$, then setting the observed reward as $R_j^{\text{obs}} \equiv R_j(A_j)$. (We use $\kappa = 0.25$ for pendigits and $\kappa =0.5$ for satimage and optdigits.) This bandit generation process makes the learning very challenging as the generated off-policy dataset has less number of observed labels. 

We randomly divide the datasets into  $70\%$ training and $30\%$ testing sets. We also randomly sequester $30\%$ of the training set as a validation set. We train all algorithms for various parameter sets on the training set, validate the hyper parameters on the validation set and test on the testing set. We evaluate our algorithm with $L_r = 2$ representation layers, and $L_p = 2$ policy layers with $50$ hidden units for representation layers and $100$ hidden units (sigmoid activation) with policy layers. We implemented/trained all algorithms in a Tensorflow environment using Adam Optimizer.

For $j$-th instance in testing data, let $h_g^{*}$ denote the optimized policy of algorithm $g$. Let $\mathcal{J}_{test}$ denote the instances in testing set and $N_{test} = |\mathcal{J}_{test}|$ denote number of instances in testing dataset. We define (absolute) accuracy of an algorithm $g$ as 
\begin{align}
\operatorname{Acc}(g) = \frac{1}{N_{test}} \sum_{j \in \mathcal{J}_{test}} \sum_{a \in \mathcal{A}} h_g^{*}(a | \boldsymbol{X}_j) R_j(a). \notag
\end{align}

We select the parameters $\lambda_1^{*} \in [0.005, 0.1], \lambda_2^{*} \in [0,0.01]$ and $\lambda_3^{*} \in [0.0001, 0.1]$ that minimize the loss given in~(\ref{eqn:cross_entropy}) estimated from the samples in the validation set. In the testing dataset, we use the full dataset to compute the accuracy of each algorithm. 

In the next subsection, we describe the performance of each algorithm on the third publicly available datasets. In each case, we run $25$ iterations, following the procedure described above; we report the average of the iterations with $95\%$ confidence intervals.

\subsubsection{Results}
In order to present a tough challenge to our algorithm we assume that the true propensities are not known and so must be estimated. Table $3$ describes the absolute accuracy of each algorithm on each dataset. As can be seen, our algorithm outperforms all the benchmarks in each dataset within $95\%$ confidence levels.  

We define loss with respect to the ``perfect'' algorithm that would predict accurately all of the time, so the {\em loss} of the algorithm $g$ is $1 -\text{Acc}(g)$.  We evaluate the improvement of our algorithm over each other algorithm as the ratio of the actual loss reduction  to the possible loss reduction, expressed as a percentage: 
$$
\text{Improvement Score}(g) = \frac{\text{Acc}({\rm PONN}{\rm -}{\rm B}) - \text{Acc}(g)}{1 - \text{Acc}(g)}
$$
The Improvement Score of each algorithm $g$ with respect to our algorithm is presented in Table $4$. Note that our algorithm achieves significant Improvement Scores in all three datasets.
 
\begin{table*}[] 
\label{table:res}
\normalsize
\centering 
\begin{tabular}{|c|c|c|c|} 
\hline 
Algorithm/Dataset   & pendigits    & satimage    & optdigits \\ \hline 
PONN-B
   & $\boldsymbol{88.01\% \pm 1.52 \%}$  & $\boldsymbol{79.22\% \pm 0.42\%}$ & $\boldsymbol{79.98\% \pm 0.62\%}$  \\ \hline 
PONN  & $85.45\% \pm 0.85 \%$  & $77.90\% \pm 0.45\%$  & $75.46\% \pm 0.57\%$  \\ \hline 
POEM-B  & $71.32\% \pm 0.73 \%$ & $45.15\% \pm 2.05\%$  & $62.14\% \pm 0.75\%$      \\ \hline 
POEM  & $68.98\% \pm 1.54 \%$ & $41.76\% \pm 2.05\%$  & $59.49\% \pm 1.53\%$      \\ \hline 
POEM-L1  & $70.84\% \pm 0.75 \%$ & $45.93\% \pm 1.01\%$ & $60.75\% \pm 0.83\%$       \\ \hline 
MLP-L1  & $83.16\% \pm 0.51\%$ & $65.95\% \pm 6.42\%$ & $75.28\% \pm 0.83\%$       \\ \hline 
LR-L1 & $80.84\% \pm 0.35\%$ & $67.45\% \pm 4.28\%$& $77.07\% \pm 0.07\%$       \\ \hline 
Logging & $10.12\% \pm 0.04\%$ & $16.55\% \pm 0.54\%$ & $10.24\% \pm 0.08\%$      \\ \hline 
\end{tabular} 
\caption{Absolute Accuracy in the  UCI Experiment (with $95\%$ CI)} 
\end{table*}

\begin{table*}[] 
\label{table:res_is}
\normalsize
\centering 
\begin{tabular}{|c|c|c|c|c|} 
\hline 
Algorithm/Dataset    & pendigits    & satimage    & optdigits \\ \hline 
PONN & $17.59\%$&  $5.52\%$& $18.41\%$       \\ \hline 
POEM-B & $58.19\%$&  $61.93\%$& $47.12\%$       \\ \hline
POEM & $61.34\%$&  $64.32\%$& $50.58\%$       \\ \hline 
POEM-L1 & $58.88\%$ & $61.56\%$ & $48.99\%$       \\ \hline 
MLP-L1& $28.80\%$  & $38.97\%$ & $53.71\%$      \\ \hline 
LR-L1 & $37.42\%$ & $36.15\%$ & $19.01\%$       \\ \hline 
Logging  & $86.65\%$ & $75.09\%$ & $77.69\%$       \\ \hline 
\end{tabular} 
\caption{Improvement scores in the  UCI Experiment} 
\end{table*}

\subsection{Chemotherapy Regimens for Breast Cancer Patients}

In this subsection, we apply our algorithm to the choice of recommendations of  chemotherapy regimen for breast cancer patients. We evaluate  our algorithm on a dataset of 10,000 records of breast cancer patients participating in the National Surgical Adjuvant Breast and Bowel Project (NSABP) by~\cite{yoon2016discovery}. Each instance consists of the following information about the patient: age, menopausal, race, estrogen receptor, progesterone receptor, human epidermal growth factor receptor 2 (HER2NEU),  tumor stage, tumor grade,  Positive Axillary Lymph Node Count(PLNC), WHO score, surgery type, Prior Chemotherapy, prior radiotherapy and histology.  The treatment is a choice among six chemotherapy regimes AC, ACT, AT, CAF, CEF, CMF. The outcomes for these regimens were derived based on 32 references from PubMed Clinical Queries. The rewards for these regimens were derived based on 32 references from PubMed Clinical Queries; this is a medically accepted procedure. The details are given in~\cite{yoon2016discovery}. 

Using these derived rewards, we construct a dataset. In this dataset, an instance is described by a triple $(\boldsymbol{X}, A, R)$, where $\boldsymbol{X}$ is the $15$-dimensional feature vector encoding the information about the particular patient, $A$ is a chemotherapy regime, and $R$ is the reward (survival/non-survival) for that chemotherapy regime for that  patient. In the dataset, \rev{$A$ is a chemotherapy regime generated in the same way as in the first experiment (with $\kappa = 0.25$)} and $R$ is the reward derived by~\cite{yoon2016discovery}.\footnote{Unfortunately, our dataset does not record which chemotherapy regime was actually chosen for each patient.}

As in the previous experiment, in comparing algorithms, we consider absolute accuracy and the improvement score. In this context, we define the absolute accuracy of an algorithm $g$ as the probability that its recommendation matches  the chemotherapy regimen with the highest reward (according to best medical practice); i.e.
$$
Acc(g) = \frac{1}{N_{test}} \sum_{j \in \mathcal{J}_{test}} \sum_{a \in \mathcal{A}} h_g^{*}(a | \boldsymbol{X}_j) \mathbb{I}(a = A^*_j)
$$ 
As before, we define the Improvement Score with respect to relative loss. 

\begin{table*}[] 
\normalsize
\centering 
\begin{tabular}{|c|c|c|} 
\hline 
Metric & Accuracy & Improvement   \\ \hline
PONN-B & $\boldsymbol{74.12\% \pm 1.25\%}$ & - \\ \hline
PONN & $62.81\% \pm 1.85\%$ & $30.41\%$ \\ \hline
POEM-B & $55.39 \% \pm 0.36\%$ & $41.98\%$  \\ \hline
POEM & $52.78 \% \pm 0.50\%$ & $45.19\%$  \\ \hline
POEM-L1 & $52.72\% \pm 0.55\%$ & $45.26\%$  \\ \hline
MLP-L1 & $61.47\% \pm 0.50\%$ & $55.05\%$ \\ \hline
LR-L1 & $51.96\% \pm 0.43\%$ & $46.12\%$  \\ \hline
Logging & $18.20\% + 1.30\%$ & $68.36\%$  \\ \hline
\end{tabular} 
\caption{Performance in the Breast Cancer Experiment} 
\label{table:res_breast}
\end{table*}

\begin{figure}
\centering
    \includegraphics[width=1\textwidth]{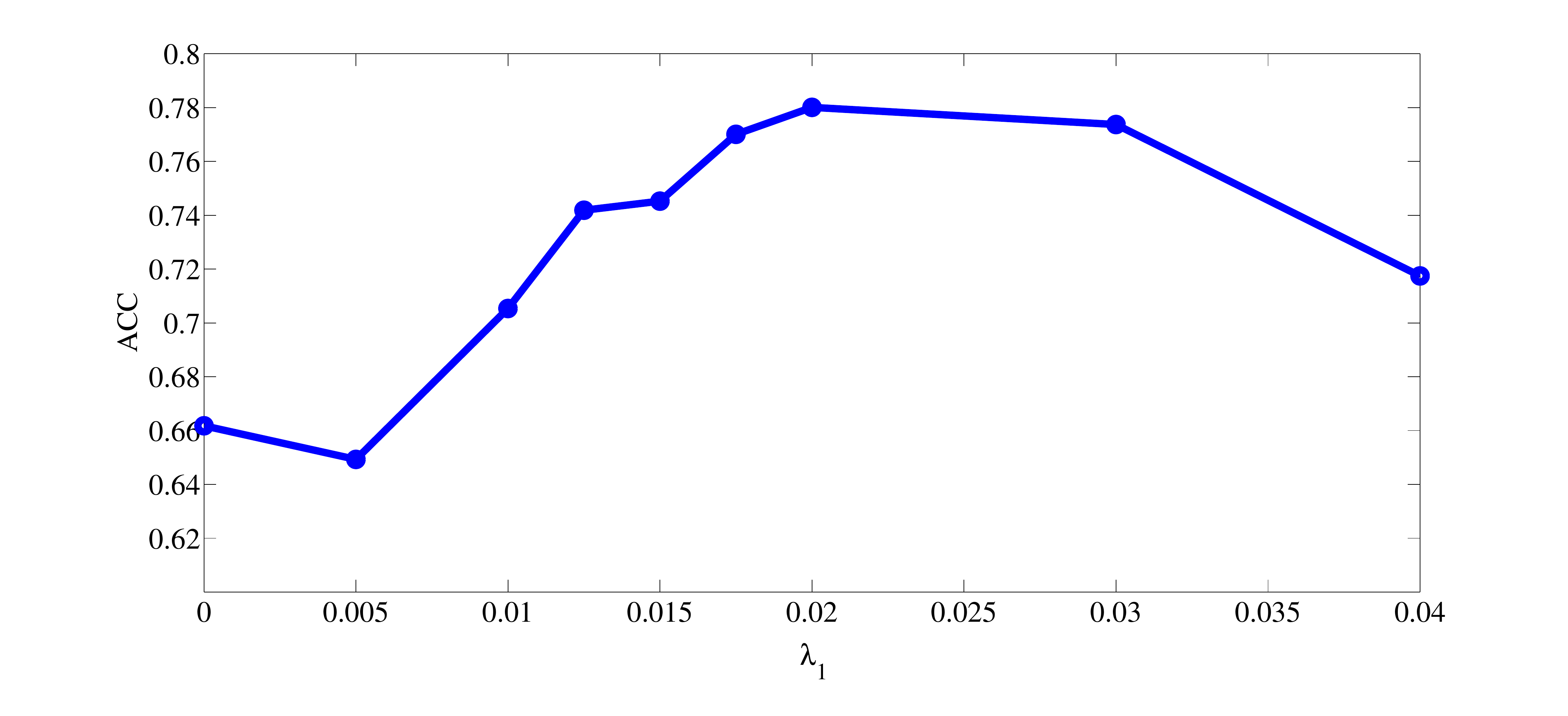}
    \caption{Effect of the hyperparameter on the accuracy of our algorithm}
\end{figure}

Table \ref{table:res_breast} describes absolute accuracy and the Improvement Scores of the our algorithm. Our algorithm achieves significant Improvement Scores with respect to all benchmarks. \rev{ There are two main reasons for these improvements. The first is that using Step B (feature selection) reduces over-fitting; this can be seen by the improvement of PONN-B over PONN and by the fact that PONN-B improves more over POEM (which does not use Step B) than over POEM-B (which does use feature selection). The second is that PONN-B allows for non-linear policies, which reduces  model misspecification. } 

\rev{Note that our action-dependent relevance discovery is also important for interpretability. The selected relevant features given by our algorithm with $\lambda_1 = 0.03$ is as follows: age, tumor stage, tumor grade for AC treatment action, age, tumor grade, lymph node status for ACT treatment action, menopausal status and surgery type for CAF treatment action, age and estrogen receptor for CEF treatment action and estrogen receptor and progesterone receptor for CMF treatment action. }

Figure 2 shows the accuracy of our algorithm for different choices of the hyper parameter 
$\lambda_1$.  As expected -- and seen in Figure 2 -- if $\lambda_1$ is too small then there is overfitting; if it is too large then a lot of relevant features are discarded.  We have chosen the value of $\lambda_1$ that maximizes accuracy.

\section{Conclusion}~ \label{sec:conc} 
This paper introduces a new approach and algorithm for the construction of effective policies when the dataset is biased and does not contain counterfactual information.  The heart of our method is the ability to identify a small number of (most) relevant features -- despite the bias and missing counterfactuals.  When tested on a wide variety of data, the algorithm we introduce achieves significant improvement over state-of-the-art methods. 

\section{Acknowledgement}
This research was funded by grants from NSF ECCS 1462245 and NSF IIP1533983.

\section*{Appendix}
\appendix
Here we collect the proofs of Theorems 1 and 2.  It is convenient to begin by recording some technical lemmas; the first two are in the literature; we give proofs for the other two.

\begin{lemma}[Theorem 1, ~\cite{audibert2009exploration}] Let $X_1, X_2, \ldots, X_n$ be i.i.d. random variables taking their values in $\left[0,b\right]$. Let $\mu = \mathbb{E}[X_1]$ be their common expected value. Consider the empirical sample mean $\bar{X}_n$ and variance $V_n$ defined respectively by
\begin{align}
\bar{X}_n = \frac{\sum_{i=1}^n X_i}{n} \text{ and } V_n = \frac{\sum_{i=1}^n (X_i - \bar{X}_n)^2}{n}. 
\end{align}
Then, for any $n \in \mathbb{N}$ and $\delta \in \left(0,1\right)$, with probability at least $1-\delta$,
\begin{align}
|\bar{X}_n - \mu| \leq \sqrt{\frac{2 V_n \log 3/\delta}{n}} + \frac{3 b \log 3/\delta}{n}.
\end{align}
\end{lemma}

\bigskip

For two probability distributions $\boldsymbol{P}$ and $\boldsymbol{Q}$ on a finite set $\mathcal{A} = \{1,2, \ldots, a\}$, let 
\begin{align}
\| \boldsymbol{P} - \boldsymbol{Q} \|_1 = \sum_{i=1}^a |\boldsymbol{P}(i) - \boldsymbol{Q}(i)|
\end{align}
denote the $L_1$ distance between $\boldsymbol{P}$ and $\boldsymbol{Q}$.

\bigskip

\begin{lemma}
~\cite{weissman2003inequalities} Let $\mathcal{A} = \{1,2, \ldots, a\}$.  Fix a probability distribution $\boldsymbol{P}$ on $\mathcal{A}$ and draw $n$ independent samples $\boldsymbol{X}^n = X_1,X_2, \ldots, X_n$ from $\mathcal{A}$ according to the distribution $\boldsymbol{P}$.  Let  $\widehat{\boldsymbol{P}} $ be the empirical distribution of $\boldsymbol{X}^n$.  Then, for all $\epsilon > 0$,
\begin{align}
\Pr(\|\boldsymbol{P} - \widehat{\boldsymbol{P}} \|_1 \geq \epsilon) \leq (2^a -2)e^{-\epsilon^2 n/2}.
\end{align}
\end{lemma}

\bigskip

The next two lemmas are auxiliary results used in the proof of Theorem 2.

\bigskip

\begin{lemma}  \label{lemma:bias_trIS} Let $\boldsymbol{P}_0 = \left[p_0(a | \boldsymbol{x})\right]$ be the actual propensities and $\widehat{\boldsymbol{P}}_0 =\left[\widehat{p}_0(a|\boldsymbol{x})\right]$ be the estimated propensities.  Assume that $\widehat{p}_0(a | \boldsymbol{x}) >0$ for all $a, \boldsymbol{x}$. The bias of the truncated IS estimator with propensities $\widehat{\boldsymbol{P}}_0$ is:
\begin{eqnarray*}
 \operatorname{bias}(\widehat{R}_m(a ; \widehat{\boldsymbol{P}}_0)) 
& =&  \sum_{j=1}^n \mathbb{E}\Bigg[ \frac{\bar{r}(a, \boldsymbol{X}_j)}{n} \bigg( \left( 1 - \frac{p_{0,j}}{\widehat{p}_{0,j}} \right) \mathbb{I}\left(\widehat{p}_{0,j} \geq m^{-1}\right)  \\ 
&&  \ \ \ \ \ \ + \ \left(1 - p_{0,j} m\right) \mathbb{I}\left(\widehat{p}_{0,j} \leq m^{-1}\right) \bigg) \Bigg]. \label{eqn:lemma_12} 
\end{eqnarray*}
\end{lemma}

\noindent {\bf Proof of Lemma 3 }
The proof is similar to~\cite{joachims2016counterfactual}. We have
\begin{eqnarray*}
 \bar{r}(a) &=& \frac{1}{n} \sum_{j=1}^n \mathbb{E}_{\boldsymbol{X}_j \sim \Pr(\mathcal{X})} \bar{r}(a, \boldsymbol{X}_j),  \notag \\ 
 \mathbb{E}(\widehat{R}_m(a; \widehat{\boldsymbol{P}}_0)) \notag 
&=&  \frac{1}{n} \sum_{j=1}^n \mathbb{E}_{(\boldsymbol{X}_j, A_j, R_j) \sim p_0} \bigg[ \min\left( \frac{\mathbb{I}(A_j = a)}{\widehat{p}_0(A_j | \boldsymbol{X}_j)}, m\right) R_j \bigg] \notag \\ 
&=& \frac{1}{n} \sum_{j=1}^n \mathbb{E}_{(\boldsymbol{X}_j, A_j) \sim p_0} \bigg[ \min\left( \frac{\mathbb{I}(A_j = a) }{\widehat{p}_0(a | \boldsymbol{X}_j)}, m\right) \bar{r}(a, \boldsymbol{X}_j) \bigg] \notag \\ 
&=& \sum_{j=1}^n \mathbb{E}_{\boldsymbol{X}_j \sim \Pr(\mathcal{X})} \bigg[ \frac{\bar{r}(a, \boldsymbol{X}_j)}{n} \min\left( \frac{1}{\widehat{p}_0(a | \boldsymbol{X}_j)}, m\right) p_0(a | \boldsymbol{X}_j) \bigg]. \notag
\end{eqnarray*}
It follows that
\begin{align}
&\operatorname{bias}(\widehat{R}_m(a ; \boldsymbol{P})) = \sum_{j=1}^n \mathbb{E}_{\boldsymbol{X}_j \sim \Pr(\mathcal{X})} \bigg[ \frac{\bar{r}(a, \boldsymbol{X}_j)}{n} \left(1 - \min\left( \frac{1}{\widehat{p}_0(a | \boldsymbol{X}_j)}, m \right) p_0(a | \boldsymbol{X}_j)\right)\bigg]. \label{eqn:plemma_11} 
\end{align}
Dividing  (\ref{eqn:plemma_11}) into the case for which $\widehat{p}_0(a | \boldsymbol{X}_j) \geq m^{-1}$ and the case for which $\widehat{p}_0(a | \boldsymbol{X}_j) < m^{-1}$ and then combining the results yields the desired conclusion. 

\bigskip

To state Lemma 4, we first define the expected relevance gain with truncated IPS reward using propensities $\widehat{\boldsymbol{P}}_0$ to be 
\begin{align}
& g_m(a,i; \widehat{\boldsymbol{P}}_0) = \mathbb{E} \left[ \left| \bar{r}_m(a,X_i;\widehat{\boldsymbol{P}}_0) - \bar{r}_m(a;\widehat{\boldsymbol{P}}_0)\right| \right] \notag
\end{align}
where
\begin{eqnarray*}
 \bar{r}_m(a; \widehat{\boldsymbol{P}}_0) &=& \mathbb{E}(\widehat{R}_m(a ; \widehat{\boldsymbol{P}}_0)) \notag  \\
&=& \mathbb{E}_{(\boldsymbol{X}, A, R) \sim p_0} \left[ \min\left( \frac{\mathbb{I}(A = a)}{p_0(A|\boldsymbol{X})}, m \right) R\right], \notag \\
\bar{r}_m(a, x_i ; \widehat{\boldsymbol{P}}_0) &=& \mathbb{E}(\widehat{R}_m(a, x_i ; \widehat{\boldsymbol{P}}_0)) \notag \\
&=& \mathbb{E}_{(\boldsymbol{X}, A, R) \sim p_0} \left[ \min\left( \frac{\mathbb{I}(A = a)}{p_0(A|\boldsymbol{X})}, m \right) R \bigg| X_i = x_i \right]. \notag
\end{eqnarray*}

\begin{lemma} We have:
$$
|g_m(a,i; \widehat{\boldsymbol{P}}_0) - g(a,i)|  \leq  B \left(\mathbb{E}\left[ \left|\operatorname{bias}(\widehat{R}_m(a, X_i; \widehat{\boldsymbol{P}}_0))\right|\right] + \left|\operatorname{bias}(\widehat{R}_m(a; \widehat{\boldsymbol{P}}_0))\right| \right). \notag
$$
\end{lemma}

\noindent {\bf Proof of Lemma 4 } This follows immediately by iterated expectations:
\begin{align}
& \bigg|\mathbb{E} \bigg( \ell \left( \mathbb{E}(\widehat{R}_m(a, X_i ; \widehat{\boldsymbol{P}}_0) ) - \mathbb{E}(\widehat{R}_m(a ; \widehat{\boldsymbol{P}}_0)) \right) 
 - \ell \left( \bar{r}(a, x_i) -\bar{r}(a) \right) \bigg)\bigg| \notag \\ 
& \ \ \ \ \ \ \leq \ B \mathbb{E} \bigg(\bigg|\mathbb{E}(\widehat{R}_m(a, X_i ; \widehat{\boldsymbol{P}}_0))  -\bar{r}(a, X_i)\bigg|\bigg) 
 + \  B |\mathbb{E}(\widehat{R}_m(a; \widehat{\boldsymbol{P}}_0)) -\bar{r}(a)|.
\end{align}

\bigskip

We now turn to the proofs of the theorems in the text.

\bigskip

\noindent {\bf Proof of Theorem 1 } Recall that the true relevance metric is  $g(a,i) = \mathbb{E}\left[|\bar{r}(a, x_i) - \bar{r}(a)| \right] = \sum_{x_i \in \mathcal{X}_i} \Pr(X_i = x_i)l(\bar{r}(a, x_i) - \bar{r}(a))$. For any action $a \in \mathcal{A}$ and $x_i \in \mathcal{X}_i$, we can bound the error between the estimated relevance metric and the relevance metric as 

\begin{eqnarray*}
 |\widehat{G}(a,i; \boldsymbol{P}_0) - g(a, i)| &=& \bigg| \sum_{x_i \in \mathcal{X}_i} \frac{N(x_i)}{n} \ell \left( \widehat{R}(a, x_i ; \boldsymbol{P}_0) - \widehat{R}(a; \boldsymbol{P}_0) \right) \notag \\ 
 && \ \  - \ \sum_{x_i \in \mathcal{X}_i} \frac{N(x_i)}{n} \ell \left( \bar{r}(a, x_i) - \bar{r}(a) \right) \notag \\ 
 && \ \ \ \   +\ \sum_{x_i \in \mathcal{X}_i} \frac{N(x_i)}{n} \ell \left( \bar{r}(a, x_i) - \bar{r}(a) \right) \notag \\ 
 && \ \ \ \ \ \ - \ \sum_{x_i \in \mathcal{X}_i} \Pr(X_i = x_i) \ell \left( \bar{r}(a, x_i) - \bar{r}(a) \right) \bigg| \notag \\ 
 & \leq&  \sum_{x_i \in \mathcal{X}_i} \frac{N(x_i)}{n} \left( \ell \left( \widehat{R}(a, x_i ; \boldsymbol{P}_0) - \widehat{R}(a ; \boldsymbol{P}_0) \right) - \ell \left( \bar{r}(a, x_i) - \bar{r}(a) \right) \right) \notag \\ 
 && \ \  + \  \sum_{x_i \in \mathcal{X}_i} \left( \frac{N(x_i)}{n} - \Pr(X_i = x_i) \right)  \ell \left( \bar{r}(a, x_i) - \bar{r}(a) \right) \notag \\ 
  & \leq&  B \sum_{x_i \in \mathcal{X}_i} \frac{N(x_i)}{n}  \left| \widehat{R}(a, x_i ; \boldsymbol{P}_0) - \bar{r}(a, x_i) \right| + B \left| \widehat{R}(a ; \boldsymbol{P}_0) - \bar{r}(a) \right| \notag \\ 
 && \ \  + \  \sum_{x_i \in \mathcal{X}_i} \left| \frac{N(x_i)}{n} - \Pr(X_i = x_i) \right|.
\end{eqnarray*}  

We bound each term separately. Applying Lemma 2, we see that with probability at least $1 - \delta$, we have
\begin{eqnarray}
 \sum_{x_i \in \mathcal{X}_i} \left|\Pr(X_i = x_i) - \frac{N(x_i)}{n}\right| &\leq& \sqrt{\frac{2 \ln 2^{b_i}/\delta}{n}} \notag \\ 
 &=&  \sqrt{\frac{2 \left(b_i \ln 2 + \ln 1/\delta \right)}{n}}. \label{eqn:thm1_1}
\end{eqnarray} 
Using Lemma 1 we see that, with probability at least $1 - \delta$, we have
\begin{align}
& \sum_{x_i \in \mathcal{X}_i} \frac{N(a, x_i)}{n} \left| \widehat{R}(a, x_i; \boldsymbol{P}_0) -\bar{r}(a, x_i) \right| \notag \\
& \;\;\;\;\; \leq \sum_{x_i \in \mathcal{X}_i} \frac{N(a, x_i)}{n} \bigg(\sqrt{\frac{2 V_n(a, x_i; \boldsymbol{P}_0) \ln 3/\delta}{N(a, x_i)}} + \frac{3 M \ln 3/\delta}{N(a, x_i)} \bigg) \notag \\ 
& \;\;\;\;\;\leq \sqrt{\frac{2 b_i  V_n(a, x_i; \boldsymbol{P}_0) \ln 3/\delta}{n}} +  \frac{3 M b_i \ln 3/\delta}{n}, \label{eqn:thm1_2}
\end{align}
where the the second inequality follows from an application of Jensen's inequality. Similarly, using Lemma 1, we see that with probability at least $1 - \delta$, we have
\begin{align}
&\left| \widehat{R}(a; \boldsymbol{P}_0) - \bar{r}(a)\right| \leq \sqrt{\frac{2 V_n(a; \boldsymbol{P}_0) \ln 3/\delta}{n}} + \frac{3 M \ln 3/\delta}{n}. \label{eqn:thm1_3}
\end{align} 
The desired result now follows by combining (\ref{eqn:thm1_1}, \ref{eqn:thm1_2} and \ref{eqn:thm1_3}). 

\bigskip

\noindent {\bf Proof of Theorem 2 }  Let 
$$
 \tilde{g}_m(a, i) = \sum_{c_i \in \mathcal{C}_{i,n}} \Pr(X_i \in c_i) \ell ( \bar{r}_m(a, c_i) - \bar{r}_m(a)). \notag
$$
Then, we can decompose the error into 
\begin{eqnarray}
 |\widehat{G}_m(a,i ; \widehat{\boldsymbol{P}}_0) - g(a,i)| &\leq& |\widehat{G}_m(a,i ; \widehat{\boldsymbol{P}}_0) - g_m(a, i ; \widehat{\boldsymbol{P}}_0)| +  |g_m(a, i ; \widehat{\boldsymbol{P}}_0) - g(a,i)| \notag \\
&\leq& |\widehat{G}_m(a,i ; \widehat{\boldsymbol{P}}_0) - \tilde{g}_m(a, i ; \widehat{\boldsymbol{P}}_0)| \notag \\ 
&& \ \ +\ |\tilde{g}_m(a, i; \widehat{\boldsymbol{P}}_0) - g_m(a, i; \widehat{\boldsymbol{P}}_0)| \notag\\ 
&&  \ \ \ \ +\  |g_m(a, i ; \widehat{\boldsymbol{P}}_0) - g(a,i)|. \label{eqn:thm_23} 
\end{eqnarray}
The first term (\ref{eqn:thm_23}) can be bounded by Theorem 1 by setting $s_n = \left \lceil{n^{1/3}}\right \rceil \leq n^{1/3}+1$, i.e., 
\begin{eqnarray*}
 |\widehat{G}_m(a,i ; \widehat{\boldsymbol{P}}_0) - \tilde{g}_m(a, i ; \widehat{\boldsymbol{P}}_0)| &\leq& \ \frac{\sqrt{4 B^2 \ln 3/\delta}}{n^{1/3}}\left( \sqrt{\bar{V}_n(a,i ; \widehat{\boldsymbol{P}}_0)} + \sqrt{V_n(a ; \widehat{\boldsymbol{P}}_0)} \right) \notag \\
 && \ \ + \ \frac{4m B \ln 3/\delta +  \sqrt{2 \ln 1/\delta + \ln 2}}{n^{2/3}}.
\end{eqnarray*}

The third term in (\ref{eqn:thm_23}) is the bias of the estimation due to estimated propensity scores and truncation, i.e., 
$$
|g_m(a, i ; \widehat{\boldsymbol{P}}_0) - g(a,i)| \notag \\
\leq B \left( \mathbb{E}\left[ \left|\operatorname{bias}(\widehat{R}_m(a, X_i); \widehat{\boldsymbol{P}}_0)\right| \right] +  \left|\operatorname{bias}(\widehat{R}_m(a); \widehat{\boldsymbol{P}}_0)\right| \right).  \notag
$$ 
We bound the second term in (\ref{eqn:thm_23}) 
\begin{eqnarray*}
 g_m(a, i; \widehat{\boldsymbol{P}}_0) &=& \mathbb{E}\left[ \ell (\bar{r}_m(a, X_i; \widehat{\boldsymbol{P}}_0) - \bar{r}_m(a; \widehat{\boldsymbol{P}}_0))\right] \notag \\
 &=& \mathbb{E}\left[ \ell (\bar{r}_m(a, X_i; \widehat{\boldsymbol{P}}_0) - \bar{r}_m(a, c_i; \widehat{\boldsymbol{P}}_0) + \bar{r}_m(a, c_i; \widehat{\boldsymbol{P}}_0) - \bar{r}_m(a; \widehat{\boldsymbol{P}}_0))\right] \notag \\ 
 &\leq& \mathbb{E}\left[ \ell \left( \frac{L}{n^{1/3}} + \bar{r}_m(a, c_i; \widehat{\boldsymbol{P}}_0) - \bar{r}_m(a; \widehat{\boldsymbol{P}}_0)\right)\right] \notag \\ 
 &\leq&  \frac{L B}{n^{1/3}} + \mathbb{E}\left[ \ell ( \bar{r}_m(a, c_i; \widehat{\boldsymbol{P}}_0) - \bar{r}_m(a; \widehat{\boldsymbol{P}}_0))\right]. 
\end{eqnarray*} 
where the first inequality follows from Assumption 3 and the second inequality follows from smoothness assumption on the loss function $l(\cdot)$, i.e.,
$$
l\left( \frac{L}{n^{1/3}} + \bar{r}_m(a, c_i; \widehat{\boldsymbol{P}}_0) - \bar{r}_m(a; \widehat{\boldsymbol{P}}_0)\right) - l\left(\bar{r}_m(a, c_i; \widehat{\boldsymbol{P}}_0) - \bar{r}_m(a; \widehat{\boldsymbol{P}}_0)\right) \leq \frac{L B}{n^{1/3}}.
$$

%\begin{acknowledgements}
%\end{acknowledgements}

\bibliographystyle{spbasic}

\end{document}